\documentclass[10pt,twocolumn,letterpaper]{article}

\usepackage{cvpr}
\usepackage{times}
\usepackage{epsfig}
\usepackage{graphicx}
\usepackage{amsmath}
\usepackage{amssymb}

\usepackage{url}            
\usepackage{booktabs}       
\usepackage{amsfonts}       
\usepackage{amsmath}
\usepackage{nicefrac}       
\usepackage{microtype}      
\usepackage{enumitem}
\usepackage{xspace}
\usepackage{xcolor} 

\newcommand*\samethanks[1][\value{footnote}]{\footnotemark[#1]}

\usepackage[pagebackref=true,breaklinks=true,letterpaper=true,colorlinks,bookmarks=false]{hyperref}

\cvprfinalcopy 

\def\cvprPaperID{****} 

\ifcvprfinal\fi
\begin{document}

\title{SoPhie: An Attentive GAN for Predicting Paths Compliant to Social and Physical Constraints}


\author{Amir Sadeghian$^1$\thanks{indicates equal contribution} \quad Vineet Kosaraju$^1$\samethanks \quad Ali Sadeghian$^2$\quad Noriaki Hirose$^1$ \\
S. Hamid Rezatofighi$^{1,3}$\quad Silvio Savarese$^1$\\
    \\
    \normalsize $^1$Stanford University \quad   $^2$ University of Florida \quad  $^3$ University of Adelaide\\
    {\tt\small amirabs@stanford.edu}
}

\maketitle

\begin{abstract}
This paper addresses the problem of path prediction for multiple interacting agents in a scene, which is a crucial step for many autonomous platforms such as self-driving cars and social robots. We present \textit{SoPhie}; an interpretable framework based on Generative Adversarial Network (GAN), which leverages two sources of information, the path history of all the agents in a scene, and the scene context information, using images of the scene. To predict a future path for an agent, both physical and social information must be leveraged. Previous work has not been successful to jointly model physical and social interactions. Our approach blends a social attention mechanism with a physical attention that helps the model to learn where to look in a large scene and extract the most salient parts of the image relevant to the path. Whereas, the social attention component aggregates information across the different agent interactions and extracts the most important trajectory information from the surrounding neighbors.
SoPhie also takes advantage of GAN to generates more realistic samples and to capture the uncertain nature of the future paths by modeling its distribution. All these mechanisms enable our approach to predict socially and physically plausible paths for the agents and to achieve state-of-the-art performance on several different trajectory forecasting benchmarks. 
\end{abstract}

\section{Introduction}
\label{sec:Introduction}

\begin{figure}[t]
  \centering
    \includegraphics[width=0.9\linewidth]{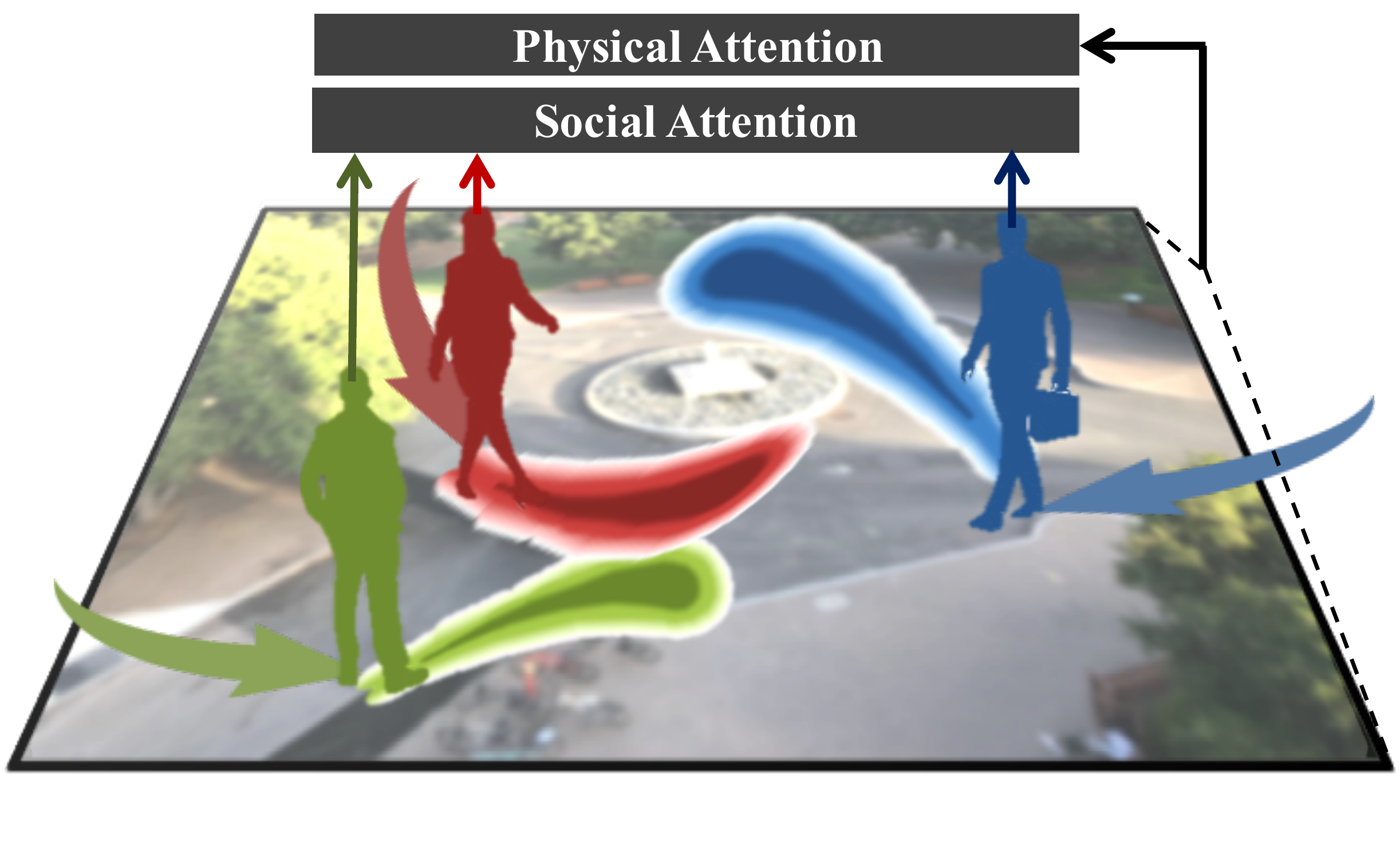}
	\caption{\small \textbf{SoPhie} predicts trajectories that are socially and physically plausible. To perform this, our approach incorporates the influence of all agents in the scene as well as the scene context.}
	\label{fig:overview}
\end{figure}

When people navigate through a park or crowded mall, they follow common sense rules in view of social decorum to adjust their paths. At the same time, they are able to adapt to the physical space and obstacles in their way. Interacting with the physical terrain as well as humans around them is by no means an easy task; because it requires:

\begin{itemize}[leftmargin=*]
\item \textbf{Obeying physical constraints of the environment.} In order to be able to walk on a feasible terrain and avoid obstacles or similar physical constraints, we have to process the local and global spatial information of our surroundings and pay attention to important elements around us. For example, when reaching a curved path, we focus more on the curve rather than other constraints in the environment, we call this \textit{physical attention}.
\item \textbf{Anticipating movements and social behavior of other people.} To avoid collisions with other people, disturbing their personal space, or interrupting some social interactions (e.g. a handshake), we must have a good understanding of others' movements and the social norms of an environment and adjust our path accordingly. We should take into account that some agents have more influence in our decision. For example, when walking in a corridor, we pay more attention to people in front of us rather than the ones behind us, we call this \textit{social attention}. Modeling these social interactions is a non-trivial task.
%
%
\item \textbf{Finding more than a single feasible path.} To get to our destination, there often exists more than a single choice for our path, which is the fuzzy nature of human motion. Indeed, there is a range for our traversable paths toward our destinations
\cite{robicquet2016learning,kitani2012activity,gupta2018social,alahi2016social}.

\end{itemize}

In this paper, we aim to tackle the problem of future path prediction for a set of agents. The existing approaches follow different strategies to solve this problem. Some methods solely rely on the scene context to predict a feasible path for each agent. For example, the approach in \cite{ballan2016knowledge} learns a dynamic pattern for all agents from patch-specific descriptors using previously created navigation maps that encode scene-specific observed motion patterns.
In~\cite{lee2017desire}, the approach learns the scene context from top-view images in order to predict future paths for each agent.  ~\cite{sadeghian2017car} applies an attention mechanism to input images in order to highlight the important regions for each agent's future path. However, all above approaches ignore the influence of the other agents' state on predicting the future path for a targeted agent. 

Parallel to path prediction using scene context information, several approaches have recently proposed to model interactions between all agents in the scene in order to predict the future trajectory for each targeted agent~\cite{fernando2017tree,fernando2017soft+}. 
Although these methods have shown promising progress in addressing this challenging problem, they still ignore the scene contexts as crucial information. In addition, these methods fall short as instead of treating pedestrian's future movements as a distribution of locations, they only predict a single path, which generally ends up optimizing ``average behavior'' rather than learning difficult constraints.. To address the second problem, \cite{alahi2016social,lee2017desire,xue2018ss} have introduced models that are able to generate multiple feasible paths. However, most of these models only incorporate the influence of few adjacent agents in a very limited search space. Recently, ~\cite{gupta2018social} proposed a GAN model that takes into account the influence of all agents in the scene.

In this work, we propose SoPhie an attentive GAN-based approach that can take into account the information from both scene context and social interactions of the agents in order to predict future paths for each agent. Influenced by the recent success of attention networks~\cite{xu2015show} and also GANs \cite{goodfellow2014generative} in different real-world problems, our proposed framework simultaneously uses both mechanisms to tackle the challenging problem of trajectory prediction. We use a visual attention model to process the static scene context alongside a novel attentive model that observes the dynamic trajectory of other agents. Then, an LSTM based GAN module is applied to learn a reliable generative model representing a distribution over a sequence of plausible and realistic paths for each agent in future.

To the best of our knowledge, no other work has previously tackled all the above problems together. SoPhie generates multiple socially-sensitive and physically-plausible trajectories and achieves state-of-the-art results on multiple trajectory forecasting benchmarks. To summarize the main contribution of the paper are as follows:
\begin{itemize}[leftmargin=*]
\item Our model uses scene context information jointly with social interactions between the agents in order to predict future paths for each agent.
\item We propose a more reliable feature extraction strategy to encode the interactions among the agents.
\item We introduce two attention mechanisms in conjunction with an LSTM based GAN to generate more accurate and interpretable socially and physically feasible paths. 
\item State-of-the-art results on multiple trajectory forecasting benchmarks.
\end{itemize}

%

\section{Related Work}
\label{sec:Related_Work}

In recent years, there have been many advances in the task of trajectory prediction. Many of the previous studies on trajectory prediction either focus on the effect of physical environment on the agents paths (agent-space interactions) and learn scene-specific features to predict future paths \cite{sadeghian2017car}, or, focus on the effect of social interactions (dynamic agent-agent phenomena) and model the behavior of agents influenced by other agents' actions \cite{alahi2016social,gupta2018social}.
Few works have been trying to combine both trajectory and scene cues \cite{lee2017desire}.

\textbf{Agent-Space Models.} This models mainly take advantage of the scene information, e.g., cars tend to drive between lanes or humans tend to avoid obstacles like benches.
Morris et al.~\cite{morris2011trajectory} cluster the spatial-temporal patterns and use hidden Markov models to model each group. Kitani et al.~\cite{kitani2012activity} use hidden variable Markov decision processes to model human-space interactions and infer walkable paths for a pedestrian. Recently, Kim et al.~\cite{kim2017probabilistic}, train a separate recurrent network, one for each future time step, to predict the location of nearby cars. 
Ballan et al.~\cite{ballan2016knowledge} introduce a dynamic Bayesian network to model motion dependencies from previously seen patterns and apply them to unseen scenes by transferring the knowledge between similar settings. 
%
In an interesting work,  a variational auto-encoders is used by Lee et al.~\cite{lee2017desire} to learn static scene context (and agents in a small neighborhood) and rank the generated trajectories accordingly.
%
Sadeghian et al.~\cite{sadeghian2017car}, also use top-view images and learn to predict trajectories based on the static scene context. Our work is similar to~\cite{sadeghian2017car} in the sense that we both use attentive recurrent neural networks to predict trajectories considering the physical surrounding; nonetheless, our model is able to take into account other surrounding agents and is able to generate multiple plausible paths using a GAN module.

\textbf{Agent-Agent Models.} Traditional models for modeling and predicting human-human interactions used ``social forces'' to capture human motion patterns~\cite{helbing1995social,mehran2009abnormal,yamaguchi2011you,pellegrini2010improving,alahi2014socially,robicquet2016learning}. The main disadvantage of these models is the need to hand-craft rules and features, limiting their ability to efficiently learn beyond abstract level and the domain experts. Modern socially-aware trajectory prediction work usually use recurrent neural networks~\cite{alahi2016social,lee2017desire,fernando2017soft+,fernando2017tree,bartoli2017context,hug2018particle}. Hug et al.~\cite{hug2017reliability} present an experiment-based study the effectiveness of some RNN models in the context socially aware trajectory prediction. This methods are relatively successful, however, most of these methods only take advantage of the local interactions and don't take into account further agents. In a more recent work, Gupta et al.~\cite{gupta2018social} address this issue as well as the fact that agent's trajectories may have multiple plausible futures, by using GANs. Nevertheless, their method treats the influence of all agents on each other uniformly. In contrast, our method uses a novel attention framework to highlight the most important agents for each targeted agent.

Few recent approaches~\cite{lee2017desire,xue2018ss,bartoli2017context}, to some extent, incorporate both the scene and social factors into their models. However, these models only consider the interaction among the limited adjacent agents and are only able to generate a single plausible path for each agent. We address all these limitations by applying wiser strategies such as 1- using visual attention component to process the scene context and highlight the most salient features of the scene for each agent, 2- using a social attention component that estimates the amount of contribution from each agent on the future path prediction of a targeted agent, and 3- using GAN to estimate a distribution over feasible paths for each agent. We support our claims by demonstrating state-of-the-art performance on several standard trajectory prediction datasets.

\section{SoPhie}
\label{sec:SoPhie}

\begin{figure*}[ht!]
  \centering
    \includegraphics[width=0.95\linewidth]{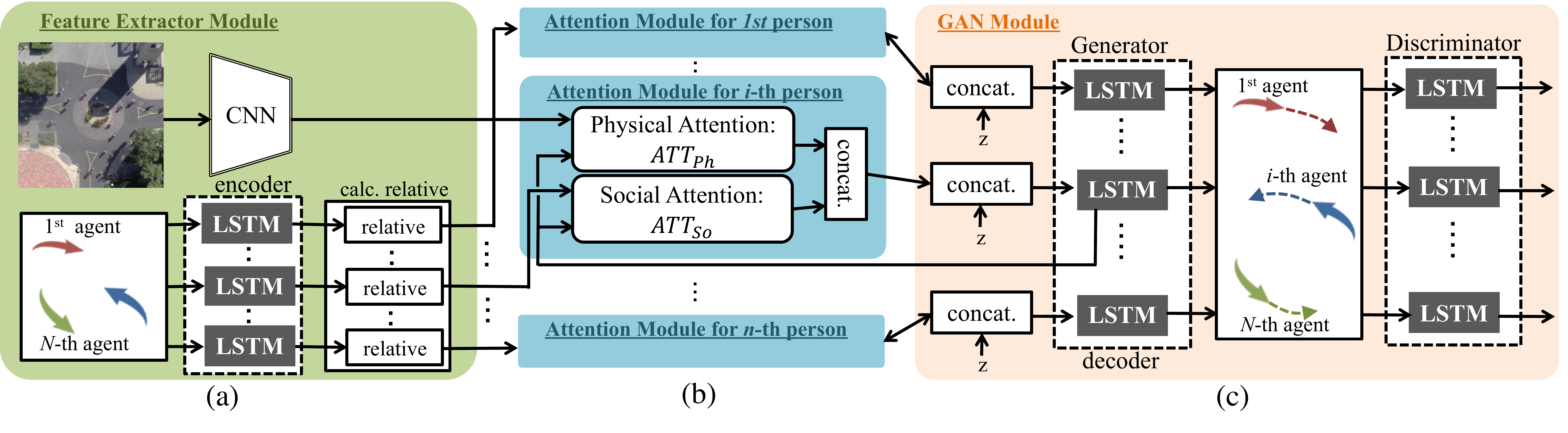}
	\caption{\small An overview of SoPhie architecture. Sophie consists of three key modules including: (a) A feature extractor module, (b) An attention module, and (c) An LSTM based GAN module.}
	\label{fig:model}
\end{figure*}

Our goal is to develop a model that can successfully predict future trajectories of a set of agents. To this end, the route taken by each agent in future needs to be influenced not only by its own state history, but also the state of other agents and physical terrain around its path. SoPhie takes all these cues into account when predicting each agent's future trajectory.

\subsection{Problem Definition}
\label{sec:Problem_Definition}
Trajectory prediction can be formally stated as the problem of estimating the state of all agents in future, given the scene information and their past states. In our case, the scene information is fed as an image $I^t$, \eg a top-view or angle-view image of the scene at time $t$, into the model. Moreover, the state of each agent $i$ at time $t$ is assumed to be its location, \eg its 2D coordinate $(x^t_i, y^t_i)\in \mathbb{R}^2$ with respect to a reference, \eg the image corner or the top view's world coordinates. Therefore, the past and current states the $N$ agents are represented by the ordered set of their 2D locations as: 
\begin{eqnarray}
\label{eq:input_def}
&&\hspace{-10mm} X^{1:t}_i = \{(x_i^\tau,y_i^\tau)| \tau = 1, \cdots, t\} \quad  \forall i\in[N],\nonumber
\end{eqnarray}
where $[N]=\{1,\cdots,N\}$. Throughout the paper, we use the notations $X^{\cdot}_{1:N}$ and $X^{\cdot}_{1:N\backslash i}$ to represent the collection of all $N$ agents' states and all agents' states excluding the target agent $i$, respectively. We also use the notation $Y^\tau$, to represent the future state in $t+\tau$. Therefore, the future ground truth and the predicted states of the agent $i$, between frames $t+1$ and $t+T$ for $T>1$, are denoted by $Y^{1:T}_i$ and $\hat{Y}^{1:T}_i$ respectively, where 
\begin{eqnarray}
\label{eq:gt_def_a}
Y^{1:T}_i  = \{(x_i^{\tau},y_i^{\tau})| \tau = t+1, \cdots, t+T  \} \quad \forall i\in[N]. \nonumber
\end{eqnarray}
Our aim is to learn the parameters of a model $W^*$ in order to predict the future states of each agent between $t+1$ and $t+T$, given the input image at time $t$ and all agents' states up to the current frame $t$, \ie 
\begin{eqnarray}
\label{eq:gt_def_b}
\hat{Y}^{1:T}_i = f(I^t, X^{1:t}_{i}, X^{1:t}_{1:N\backslash i};W^*), \nonumber
\end{eqnarray}
where the model parameters $W^*$ is the collection of the weights for all deep neural structures used in our model. We train all the weights end-to-end using back-propagation and stochastic gradient descent by minimizing a loss $\mathcal{L}_{GAN}$ between the predicted and ground truth future states for all agents. We elaborate the details in the following section.

\subsection{Overall Model}
\label{sec:Overall_Model}
Our model consists of three key components including: 1- A feature extractor module, 2- An attention module, and 3- An LSTM based GAN module (Fig. \ref{fig:model}). First, the feature extractor module extracts proper features from the scene, \ie the image at the current frame $I^t$, using a convolutional neural network. It also uses an LSTM encoder to encode an index invariant, but temporally dependent, feature between the state of each agent, $X^{1:t}_{i}$, and the states of all other agents up to the current frame, $X^{1:t}_{1:N\backslash i}$ (Fig. \ref{fig:model}(a)). Then, the attention module highlights the most important information of the inputted features for the next module (Fig. \ref{fig:model} (b)). The attention module consists of two attention mechanisms named as \emph{social} and \emph{physical} attention components. The physical attention learns the spatial (physical) constraints in the scene from the training data and concentrates on physically feasible future paths for each agent. Similarly, the social attention module learns the interactions between agents and their influence on each agent's future path. Finally, the LSTM based GAN module (Fig. \ref{fig:model} (c)) takes the highlighted features from the attention module to generate a sequence of plausible and realistic future paths for each agent. In more details, an LSTM decoder is used to predict the temporally dependent state of each agent in future, \ie $\hat{Y}^{1:T}_i$. Similar to GAN, a discriminator is also applied to improve the performance of the generator model by forcing it to produce more realistic samples (trajectories). In the following sections, we elaborate each module in detail.

\subsection{Feature extractors}

The feature extractor module has two major components, explained below.
To extract the visual features $V^t_{Ph}$ from the image $I^t$, we use a Convolutional Neural Network (CNN).
\begin{eqnarray}
\label{eq:Feature_extractors1}
V^t_{Ph} = CNN(I^t ; W_{cnn}) 
\end{eqnarray}
In this paper, we use VGGnet-19 \cite{simonyan2014very} as $CNN(\cdot)$, where its weights $W_{cnn}$ is initialized by pre-training on ImageNet \cite{russakovsky2015imagenet} and fine-tuning on the task of scene segmentation as described in \cite{long2015fully}.

To extract joint features from the past trajectory of all agents, we perform the following procedure. Similar to ~\cite{gupta2018social}, first an LSTM is used to capture the temporal dependency between all states of an agent $i$ and encode them into a high dimensional feature representation for time $t$, \ie
\begin{eqnarray}
\label{eq:Feature_extractors2}
V^t_{en}(i) = LSTM_{en}(X^{t}_{i}, h_{en}^{t}(i); W_{en}),
\end{eqnarray}
where $h_{en}^{t}(i)$ represents the hidden state of the encoder LSTM at time $t$ for the agent $i$.
Moreover, to capture the influence of the other agents' state on the prediction of the future trajectory of an agent, we need to extract a joint feature from all agents' encoded features $V^t_{en}(\cdot)$. However, this joint feature cannot be simply created by concatenating them as the order of the agents does matter. To make the joint feature permutation invariant with respect to the index of the agents, the existing approaches use a permutation invariant (symmetric) function such as \emph{max}~\cite{gupta2018social}. Then, this joint global feature is concatenated by each agent's feature $V^t_{en}(i)$ to be fed to the state generator module. However this way, all agents will have an identical joint feature representation. In addition, the permutation invariant functions such as \emph{max} may discard important information of their inputs as they might loose their uniqueness. To address these two limitations, we instead define a consistent ordering structure, where the joint feature for a target agent $i$ is constructed by sorting the other agents' distances from agent $i$, \ie
\begin{eqnarray}
\label{eq:Feature_extractors3}
V^t_{So}(i) = \big(V^t_{en}(\pi_j) - V^t_{en}(i) \big| \forall \pi_j\in[N]\backslash i)\big),
\end{eqnarray}
where $\pi_j$ is the index of the other agents sorted according to their distances to the target agent $i$. In this framework, each agent $i$ has its own unique joint (social) feature vector. We also use \emph{sort} as the permutation invariant function, where the reference for ordering is the euclidean distance between the target agent $i$ and other agents. Note that \emph{sort} function is advantageous in comparison with \emph{max} as it can keep the uniqueness of the input.    
To deal with variable number of agents, we set a maximum number of agents ($N = N_{max}$) and use a dummy value as features if the corresponding agent does not exist in the current frame. 



\subsection{Attention Modules}
\label{sec:Attention_Modules}
Similar to humans who pays more attention to close obstacles, upcoming turns and people walking towards them, than to the buildings or people behind them, we want the model to focus more on the salient regions of the scene and the more relevant agents in order to predict the future state of each agent. To achieve this, we use two separate soft attention modules similar to \cite{xu2015show} for both physical $V^t_{Ph}$ and social $V^t_{So}(i)$ features.
	
\textbf{Physical Attention} The inputs to this attention module $ATT_{Ph}(\cdot)$ are the hidden states of the decoder LSTM in the GAN module, and the visual features extracted from the image $V^t_{Ph}$. Note that, the hidden state of the decoder LSTM has the information for predicting the agent's future path. And this module learns the spatial (physical) constraints in the scene from the training data. Therefore, the output would be a context vector $C^t_{Ph}$, which concentrates on feasible paths for each agent. 
\begin{eqnarray}
\label{eq:pcontext}
C^t_{Ph}(i) = ATT_{Ph}(V^t_{Ph}, h_{dec}^{t}(i); W_{Ph})
\end{eqnarray}
Here, $W_{Ph}$ are the parameters of the physical attention module and $h^{t}_{dec}(i)$ represents the hidden state of the decoder LSTM at time $t$ for the agent $i$.


\textbf{Social Attention} Similar to the physical attention module, the joint feature vector $V^t_{So}(i)$ together with the hidden state of the decoder LSTM for the $i$-th agent, are fed to the social attention module $ATT_{So}(\cdot)$ with the parameters $W_{So}$ to obtain a social context vector $C^t_{So}(i)$ for the $i$-th agent. This vector highlights which other agents are most important to focus on when predicting the trajectory of the agent $i$.
\begin{eqnarray}
\label{eq:scontext}
C^t_{So}(i) = ATT_{So}(V^{t}_{So}(i), h^{t}_{dec}(i); W_{So})
\end{eqnarray}

We use soft attention similar to \cite{xu2015show} for both $ATT_{Ph}(\cdot)$ and $ATT_{So}(\cdot)$, which is differentiable and the whole architecture can be trained end-to-end with back-propagation. Social attention and physical attention aggregate information across all the involved agents and the physical terrain to deal with the complexity of modeling the interactions of all agents in crowded areas while adding interpretability to our predictions. This also suppresses the redundancies of the input data in a helpful fashion, allowing the predictive model to focus on the important features. Our experiments show the contribution of our attention components in Table \ref{quantitative-table-one}.

\subsection{LSTM based Generative Adversarial Network}
\label{sec:Generator}
In this section, we present our LSTM based Generative Adversarial Network (GAN) module that takes the social and physical context vectors for each agent $i$, $C^t_{So}(i)$ and $C^t_{Ph}(i)$, as input and outputs candidate future states which are compliant to social and physical constraints. Most existing trajectory prediction approaches use the L2 norm loss between the ground truth and the predictions to estimate the future states \cite{sadeghian2017car}. By using L2 loss, the network only learns to predict one future path for each agent, which is intuitively the average of all feasible future paths for each agent. Instead, in our model, we use GAN to learn and predict a distribution over all the feasible future paths.

GANs consist of two networks, a \emph{generator} and a \emph{discriminator} that compete with each other. The generator is trained to learn the distribution of the paths and to generate a sample of the possible future path for an agent while the discriminator learns to distinguish the feasibility or infeasibility of the generated path. These networks are simultaneously trained in a two player min-max game framework. 
In this paper similar to~\cite{gupta2018social}, we use two LSTMs, a decoder LSTM as the generator and a classifier LSTM as the discriminator, to estimate the temporally dependent future states.

%
%

\textbf{Generator (G)} Our generator is a decoder LSTM, $LSTM_{dec}(\cdot)$. Similar to the conditional GAN~\cite{mirza2014conditional}, the input to our generator is a white noise vector $z$ sampled from a multivariate normal distribution while the physical and social context vectors are its conditions. We simply concatenate the noise vector $z$ and these context vectors as the input, \ie $C^{t}_G(i) = [C^t_{So}(i), C^t_{Ph}(i), z]$. Thus, the generated $\tau^{th}$ future state's sample for each agent is attained by:
\begin{eqnarray}
\label{eq:gen_dec}
& \hat{Y}_i^{\tau} = LSTM_{dec}\big(C^{t}_G(i), h^{\tau}_{dec}(i); W_{dec}\big), 
\end{eqnarray}


%


%
\textbf{Discriminator (D)}
The discriminator in our case is another LSTM, $LSTM_{dis}(\cdot)$, which its input is a randomly chosen trajectory sample from the either ground truth or predicted future paths for each agent up to $\tau^{th}$ future time frame, \ie $T_i^{1:\tau}\sim p(\hat{Y}_i^{1:\tau},Y_i^{1:\tau} )$ 
\begin{eqnarray}
\label{eq:dis}
\hat{L}_{i}^{\tau} = LSTM_{dis}(T_i^{\tau}, h_{dis}^{\tau}(i); W_{dis}), 
\end{eqnarray}
%
where $\hat{L}_{i}^{\tau}$ is the predicted label from the discriminator for the chosen trajectory sample to be a ground truth (real) $Y_i^{1:\tau}$ or predicted (fake) $\hat{Y}_i^{1:\tau}$ with the truth label $L_{i}^{\tau} = 1$ and $L_{i}^{\tau} = 0$,  respectively. The discriminator forces the generator to generate more realistic (plausible) states. 

\textbf{Losses}
To train SoPhie, we use the following losses:
\begin{eqnarray}
\label{eq:obj}
W^* =\operatorname*{argmin}_W \quad\mathbb{E}_{i,\tau}[\mathcal{L}_{GAN}\big(\hat{L}_{i}^{\tau}, L_{i}^{\tau} \big)+ \nonumber\\
\lambda \mathcal{L}_{L2}(\hat{Y}_i^{1:\tau},Y_i^{1:\tau})],
\end{eqnarray}
where $W$ is the collection of the weights of all networks used in our model and $\lambda$ is a regularizer between two losses. 

The adversarial loss $\mathcal{L}_{GAN}(\cdot,\cdot)$ and L2 loss $\mathcal{L}_{L2}(\cdot,\cdot)$ are shown as follows:\\
$\mathcal{L}_{GAN}\big(\hat{L}_{i}^{\tau}, L_{i}^{\tau} \big) = $
\begin{eqnarray}
\label{eq:loss1}
\qquad \min_{G} \max_{D}\quad \mathbb{E}_{T^{1:\tau}_i\sim p(Y^{1:\tau}_i)}[L^{\tau}_i\mbox{log} \hat{L}^{\tau}_i] +\nonumber\\\quad
\mathbb{E}_{T^{1:\tau}_i\sim p(\hat{Y}^{1:\tau}_i)}[(1-L^{\tau}_i)\mbox{log} (1 - \hat{L}^{\tau}_i)], 
\end{eqnarray}
\begin{eqnarray}
\label{eq:loss2}
\mathcal{L}_{L2}(\hat{Y}_i^{\tau},Y_i^{\tau}) = ||\hat{Y}_i^{\tau}-Y_i^{\tau}||^2_2.
\end{eqnarray}
%

\section{Experiments}
\label{sec:Experiments}

In this section, we first evaluate our method on the commonly used datasets such as ETH \cite{pellegrini2010improving} and UCY \cite{lerner2007crowds}, and on a recent and larger dataset, \ie Stanford drone dataset \cite{robicquet2016learning}. We also compare its performance against the various baselines on these datasets. Next, we present a qualitative analysis of our model on the effectiveness of the attention mechanisms. Finally, we finish the section by demonstrating some qualitative results on how our GAN based approach provides a good indication of path traversability for agents.

\begin{table*}[t!] 
  \centering
  \resizebox{0.9\linewidth}{!}{%
  \begin{tabular}{l|l|l|l|l|l|l|l|l|l|l}
    \midrule
    & \multicolumn{5}{c|}{\textbf{Baselines}} & \multicolumn{5}{c}{\textbf{SoPhie (Ours)}} \\
    \cmidrule[1pt]{2-11}
    \textbf{Dataset} & \textbf{Lin} & \textbf{LSTM} & \textbf{S-LSTM} & \textbf{S-GAN} & \textbf{S-GAN-P} & $\mathbf{T_{A}}$ & $\mathbf{T_O+I_O}$ & $\mathbf{T_O+I_A}$ & $\mathbf{T_A+I_O}$ & $\mathbf{T_A+I_A}$ \\
    \midrule
\textbf{ETH} & 1.33 / 2.94 & 1.09 / 2.41 & 1.09 / 2.35 & 0.81 / 1.52  & 0.87 /  1.62 & 0.90 / 1.60 & 0.86 / 1.65 & 0.71 / 1.47 & 0.76 / 1.54 & \textbf{0.70} / \textbf{1.43}     \\
    \textbf{HOTEL} & \textbf{0.39} / \textbf{0.72} & 0.86 / 1.91 & 0.79 / 1.76 &  0.72 / 1.61 & 0.67 / 1.37 & 0.87 / 1.82 & 0.84 / 1.80 & 0.80 / 1.78 & 0.83 / 1.79 & 0.76 / 1.67      \\
    \textbf{UNIV}  & 0.82 /  1.59 &  0.61 /  1.31 & 0.67 / 1.40 & 0.60 / 1.26 & 0.76 / 1.52 & \textbf{0.49} / \textbf{1.19} & 0.58 / 1.27 & 0.55 / 1.23 & 0.55 / 1.25 & 0.54 / 1.24  \\
    \textbf{ZARA1} & 0.62 / 1.21 & 0.41 / 0.88 & 0.47 / 1.00 & 0.34 / 0.69 & 0.35 / 0.68 & 0.38 / 0.72 & 0.34 / 0.68 & 0.35 / 0.67 & 0.32 / 0.64 & \textbf{0.30} / \textbf{0.63}  \\
    \textbf{ZARA2} & 0.77 / 1.48& 0.52 / 1.11 & 0.56 /  1.17&  0.42 / 0.84 & 0.42 / 0.84 & 0.38 / 0.79 & 0.40 / 0.82 & 0.43 / 0.87 & 0.41 / 0.80 & \textbf{0.38} / \textbf{0.78}  \\
    \toprule
    \midrule
    \textbf{AVG}   & 0.79 / 1.59 & 0.70 /  1.52  & 0.72 / 1.54 & 0.58 / 1.18 & 0.61 / 1.21 & 0.61 / 1.22 & 0.61 / 1.24 & 0.57 / 1.20 & 0.58 / 1.20 & \textbf{0.54} / \textbf{1.15}  \\
    \toprule
  \end{tabular}
  }
  \caption{\small Quantitative results of baseline models vs. SoPhie architectures across datasets on the task of predicting 12 future timesteps, given the 8 previous ones. Error metrics reported are ADE / FDE in meters. SoPhie models consistently outperform the baselines, due to the combination of social and physical attention applied in a generative model setting.}
  \label{quantitative-table-one}
\end{table*}

\begin{table*}[t!]
  \centering
  \resizebox{1\linewidth}{!}{ 
  \begin{tabular}{l|l|l|l|l|l|l|l|l|l|l|l}
    \midrule
    & \multicolumn{6}{c|}{\textbf{Baselines}} & \multicolumn{5}{c}{\textbf{SoPhie (Ours)}} \\
    \cmidrule[1pt]{2-12}
    \textbf{Dataset} & \textbf{Lin} & \textbf{SF} & \textbf{S-LSTM} & \textbf{S-GAN} & \textbf{CAR-Net} & \textbf{DESIRE} & $\mathbf{T_{A}}$ & $\mathbf{T_O+I_O}$ & $\mathbf{T_O+I_A}$ & $\mathbf{T_A+I}$ & $\mathbf{T_A+I_A}$ \\
    \midrule
\textbf{SDD} & 37.11 / 63.51 & 36.48 / 58.14 & 31.19 / 56.97  & 27.246 / 41.440 & 25.72 / 51.8  & 19.25 / 34.05 & 17.76 / 32.14 & 18.40 / 33.78 & 16.52 / 29.64 & 17.57 / 33.31 & \textbf{16.27} / \textbf{29.38}  \\
    \toprule
  \end{tabular}
  }
  \caption{\small ADE and FDE in pixels of various models on Stanford Drone Dataset. SoPhie's main performance gain comes from the joint introduction of social and physical attention applied in a generative modeling setting.}
  \label{quantitative-table-two}
\end{table*}

\textbf{Datasets} We perform baseline comparisons and ablation experiments on three core datasets. First, we explore the publicly available ETH \cite{pellegrini2010improving} and UCY \cite{lerner2007crowds} datasets, which both contain annotated trajectories of real world pedestrians interacting in a variety of social situations. These datasets include nontrivial movements including pedestrian collisions, collision avoidance behavior, and group movement. Both of the datasets consists of a total of five unique scenes, Zara1, Zara2, and Univ (from UCY), and ETH and Hotel (from ETH). Each scene includes top-view images and 2D locations of each person with respect to the world coordinates. One image is used per scene as the cameras remain static. Each scene occurs in a relatively unconstrained outdoor environment, reducing the impact of physical constraints. 
We also explore the Stanford Drone Dataset (SDD) \cite{robicquet2016learning}, a benchmark dataset for trajectory prediction problems. The dataset consists of a bird's-eye view of 20 unique scenes in which pedestrians, bikes, and cars navigate on a university campus. Similar to the previous datasets, images are provided from a top-view angle, but coordinates are provided in pixels. These scenes are outdoors and contain physical landmarks such as buildings and roundabouts that pedestrians avoid.

\textbf{Implementation details} We iteratively trained the generator and discriminator models with the Adam optimizer, using a mini-batch size of 64 and a learning rate of 0.001 for both the generator and the discriminator. Models were trained for 200 epochs. The encoder encodes trajectories using a single layer MLP with an embedding dimension of 16. In the generator this is fed into a LSTM with a hidden dimension of 32; in the discriminator, the same occurs but with a dimension of 64. The decoder of the generator uses a single layer MLP with an embedding dimension of 16 to encoder agent positions and uses a LSTM with a hidden dimension of 32. In the social attention module, attention weights are retrieved by passing the encoder output and decoder context through multiple MLP layers of sizes 64, 128, 64, and 1, with interspersed ReLu activations. The final layer is passed through a Softmax layer. The interactions of the surrounding $N_{max}=32$ agents are considered; this value was chosen as no scenes in either dataset exceeded this number of total active agents in any given timestep. If there are less than $N_{max}$ agents, the dummy value of 0 is used. The physical attention module takes raw VGG features (512 channels), projects those using a convolutional layer, and embeds those using a single MLP to an embedding dimension of 16. The discriminator does not use the attention modules or the decoder network. When training we assume we have observed eight timesteps of an agent and are attempting to predict the next $T=12$ timesteps. We weight our loss function by setting $\lambda=1$.

/
In addition, to make our model more robust to scene orientation, we augmented the training data by flipping and rotating the scene and also normalization of agents' coordinates. We observed that these augmentations are conducive to make the trained model general enough in order to perform well on the unseen cases in the test examples and different scene geometries such as roundabouts.

\textbf{Baselines \& Evaluation} For the first two datasets, a few simple, but strong, baselines are used. These include \textit{Lin}, a linear regressor that estimates linear parameters by minimizing the least square error; \textit{S-LSTM}, a prediction model that combines LSTMs with a social pooling layer, as proposed by Alahi \textit{et. al.} \cite{alahi2016social}; \textit{S-GAN} and \textit{S-GAN-P}, predictive models that applies generative modeling to social LSTMs \cite{gupta2018social}. For the drone dataset, we compare to the same linear and Social LSTM baselines, but also explore several other state-of-the-art methods. These include \textit{Social Forces}, an implementation of the same Social Forces model from \cite{yamaguchi2011you}; \textit{DESIRE}, an inverse optimal control (IOC) model proposed by Lee \textit{et. al.} that utilizes generative modeling; and \textit{CAR-Net}, a physically attentive model from \cite{sadeghian2017car}. 
For all datasets, we also present results of various versions of our SoPhie model in an ablative setting by 
1- $\mathrm{T_A}$: Sophie model with social features only and the social attention mechanism, 2- $\mathrm{T_O+I_O}$ Sophie model with both visual and social features without any attention mechanism, 3- $\mathrm{T_O+I_A}$ Sophie model with both visual and social features with only visual attention mechanism, 4- $\mathrm{T_A+I_O}$ Sophie model with both visual and social features with only social attention mechanism, and 5- $\mathrm{T_A+I_A}$ complete Sophie model with all modules. 


All models are evaluated using the average displacement error (ADE) metric defined as the average L2 distance between the ground truth and pedestrian trajectories, over all pedestrians and all time steps, as well as the final displacement error metric (FDE). The evaluation task is defined to be performed over 8 seconds, using the past 8 positions consisting of the first 3.2 seconds as input, and predicting the remaining 12 future positions of the last 4.8 seconds. For the first two datasets, we follow a similar evaluation methodology to \cite{gupta2018social} by performing a leave-one-out cross-validation policy where we train on four scenes, and test on the remaining one. These two datasets are evaluated in meter space. For the SDD, we utilize the standard split, and for the sake of comparison to baselines we report results in pixel space, after converting from meters.

\subsection{Quantitative Results}
\textbf{~~~~~ETH and UCY} We compare our model to various baselines in Table \ref{quantitative-table-one}, reporting the average displacement error (ADE) in meter space, as well as the final displacement error (FDE). As expected, we see that in general the linear model performs the worst, as it is unable to model the complex social interactions between different humans and the interactions between humans and their physical space. 
We also notice that S-LSTM provides an improvement over the linear baseline, due to its use of social pooling, and that S-GAN provides an improvement to this LSTM baseline, by approaching the problem from a generative standpoint.

Our first model, $\mathrm{T_A}$, which solely applies social context to pedestrian trajectories, performs slightly better than the S-GAN on average due to better feature extraction strategy and attention module. As expected, although social context helps the model form better predictions, it alone is not enough to truly understand the interactions in a scene. Similarly, while our second model $\mathrm{T_O+I_O}$ applies both pedestrian trajectories and features from the physical scene (no attention), the lack of any context about these additional features make the model unable to learn which components are most important, giving it a similar accuracy to $\mathrm{T_A}$. Our first major gains in model performance come when exploring the $\mathrm{T_O+I_A}$ and $\mathrm{T_A+I_O}$ models. Because the former applies physical context to image features and the latter applies social context to trajectory features, each model is able to learn the important aspects of interactions, allowing them to slightly outperform the previous models. Interestingly, $\mathrm{T_O+I_A}$ performs slightly better than $\mathrm{T_A+I_O}$ potentially suggesting that understanding physical context is slightly more helpful in a prediction task. The final SoPhie model, consisting of social attention on trajectories and physical attention on image features ($\mathrm{T_A+I_A}$) outperformed the previous models, suggesting that combining both forms of attention allows for robust model predictions.\\

\textbf{Stanford Drone Dataset} We next compare our method to various baselines in Table \ref{quantitative-table-two}, reporting the ADE and FDE in pixel space. Much like the previous datasets, with SDD we see that the linear baseline performs the worst, with S-LSTM and S-GAN providing an improvement in accuracy. The next major improvement in accuracy is made with CAR-Net, due to the use of physical attention. This is likely due to the nature of SDD, where pedestrian movements based on the curvature of the road can be extrapolated from the birds eye view of the scene. The next major improvement in accuracy is made with the DESIRE framework, which explores trajectory prediction from a generative standpoint, making it the best baseline. Note that the DESIRE results are linearly interpolated from the 4.0s result reported in \cite{lee2017desire} to 4.8s, as their code is not publicly available. 
Finally, incorporating social context in $\mathrm{T_A}$, as well as both social and physical context in $\mathrm{T_A+I_A}$ allow for significant model improvements, suggesting that both attentive models are crucial to tackling the trajectory prediction problem.\\

\textbf{Impact of social and physical constraints.} Since the goal is to produce socially acceptable paths we also used a different evaluation metrics that reflect the percentage of near-collisions (if two pedestrians get closer than the threshold of $0.10m$). We have calculated the average percentage of pedestrian near collisions across all frames in each of the BIWI/ETH scenes. These results are presented in Table 3. To better understand our model's ability to also produce physically plausible paths, we also split the test set of the Stanford Drone Dataset into two subsets: simple and complex, as previously done in CAR-Net ~\cite{sadeghian2017car} and report results in Table 4. We note that the S-GAN baseline achieves decent performance on simple scenes, but is unable to generalize well to physically complex ones. On the other hand, CAR-Net and SoPhie both achieves a slight performance increase on simple scenes over S-GAN and trajectory only LSTM, as well as nearly halving the error on complex scenes, due to this physical context. This experiment demonstrates that Sophie's use of physical attention successfully allows it to predict both physical and socially acceptable paths.

\begin{figure*}[ht!]
  \centering
    \includegraphics[width=0.98\linewidth]{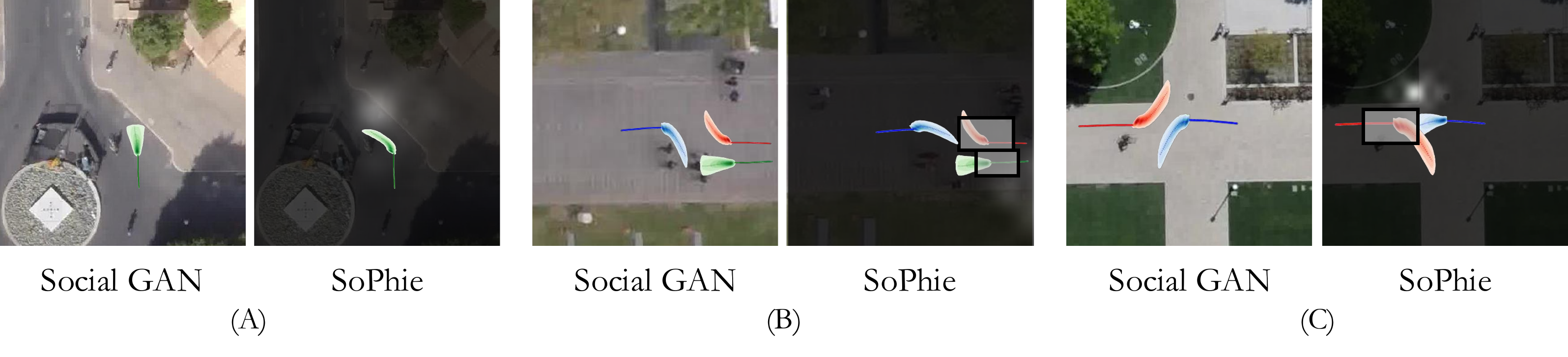}
\caption{\small Three sample scenarios where physical and social attention allow correct predictions and fixes the Social GAN errors. In all figures, past and predicted trajectories are plotted as line and distributions, respectively. We display the weight maps of the physical attention mechanism highlighted in white on the image. The white boxes on the agents show the social attention on the agents with respect to the blue agent. The size of the boxes are relative to the attention weights on different agents.
} 
	\label{fig:qual1}
\end{figure*}

\begin{figure*}[ht!]
  \centering
    \includegraphics[width=0.98\linewidth]{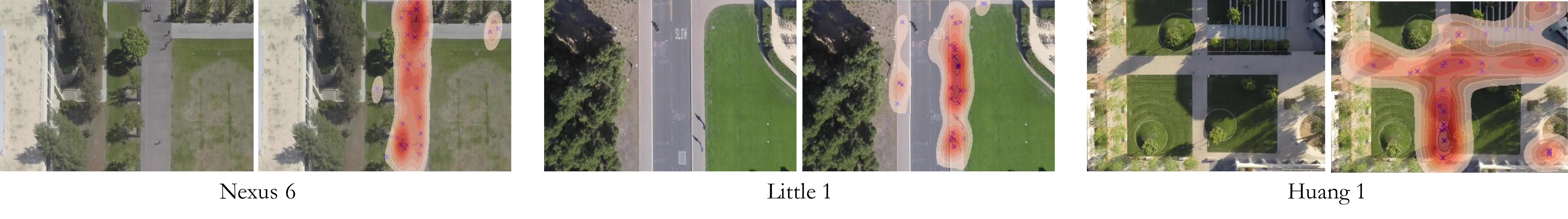}  
	\caption{\small Using the generator to sample trajectories and the discriminator to validate those paths, we present highly accurate traversability maps for SDD scenes. Maps are presented in red, and generated only with 30 starting samples, illustrated as blue crosses.} 
	\label{fig:qual2}
\end{figure*}

\begin{table}[ht!]
\centering
\begin{tabular}{l|l|l|l|l} 
               & \textbf{GT} & \textbf{LIN}   & \textbf{S-GAN} & \textbf{SoPhie} \\ \hline
\textbf{ETH}   & 0.000       & 3.137          & 2.509          & \textbf{1.757}  \\
\textbf{HOTEL} & 0.092       & \textbf{1.568} & 1.752          & 1.936           \\
\textbf{UNIV}  & 0.124       & 1.242          & \textbf{0.559} & 0.621           \\
\textbf{ZARA1} & 0.000       & 3.776          & 1.749          & \textbf{1.027}  \\
\textbf{ZARA2} & 0.732       & 3.631          & 2.020          & \textbf{1.464}  \\\hline
\textbf{Avg} & 0.189       & 2.670          & 1.717          & \textbf{1.361} \\\bottomrule
\end{tabular}
\caption{\small{Average \% of colliding pedestrians per frame for each of the scenes in BIWI/ETH. A collision is detected if the euclidean distance between two pedestrians is less than 0.10m.}}
\label{tab:collision}
\end{table}

\begin{table}[ht!]
\centering
  \resizebox{0.6\linewidth}{!}{ 

\begin{tabular}{l|l|l} \hline 
Model               & Complex & Simple \\ \hline
\textbf{LSTM}   & 31.31       & 30.48    \\
\textbf{CAR-Net} & 24.32       & 30.92    \\
\textbf{S-GAN}  & 29.29       & \textbf{22.24}      \\
\textbf{SoPhie} & \textbf{15.61}       & \textbf{21.08}   \\ \hline
\end{tabular}
}
\caption{\small{Performance of multiple baselines on the Stanford Drone Dataset, split into physically simple and complex scenes. Error is ADE and is reported in pixels.}}
\label{tab:collision}
\end{table}

\subsection{Qualitative Results}

We further investigate the ability of our architecture to model how social and physical interactions impact future trajectories. Fig. \ref{fig:qual1} demonstrates the affects that attention can have in correcting erroneous predictions. Here we visualize three unique scenarios, comparing a baseline Social GAN prediction to that of our model. In the first scenario (A), physical attention ensures the trajectory of the green pedestrian follows the curve of the road. In the second, scenario B, social attention on the green pedestrian ensures that the main blue pedestrian does not collide with either pedestrian. In the third scenario (C), physical attention is applied to ensure the red pedestrian stays within the road, while social attention ensures that the blue pedestrian does not collide with the red one. As such, the introduction of social and physical attention not only allows for greater model interpretability but also better aligns predictions to scene constraints. 

An additional benefit of the generative SoPhie architecture is that it can be used to understand which areas in a scene are traversable.
To show the effectiveness of our method, we sampled 30 random agents from the test set (i.e., first 8 seconds of each trajectory)
and the generator generated sample trajectories using this starting points. These generated trajectories were then validated using the discriminator. The distribution of these trajectories results in an interpretable traversability map, as in Fig. \ref{fig:qual2}. Each image represents a unique scene from SDD, with the overlayed heatmap showing traversable areas and the blue crosses showing the starting samples. With Nexus 6, the model is able to successfully identify the traversable areas as the central road and the path to the side. With Little 1, the model identifies the main sidewalk that pedestrians walk on while correctly ignoring the road that pedestrians avoid. In Huang 1, the model is able to correctly identify the cross section as well as side paths on the image. We thus observe that the generative network can successfully be used to explore regions of traversability in scenes even with a small number of samples. 


\section{Conclusion}
We propose a trajectory prediction framework that outperforms state-of-the-art methods on multiple benchmark datasets. Our method leverages complete scene context and interactions of all agents, while enabling interpretable predictions, using social and physical attention mechanisms. To generate a distribution over the predicted trajectories, we proposed an attentive GAN which can successfully generate multiple physically acceptable paths that respect social constraints of the environment. We showed that by modeling jointly the information about the physical environment and interactions between all agents, our model learns to perform better than when this information is used independently.

\pagebreak

{\small
\bibliographystyle{ieee}
\bibliography{egbib}
}

\end{document}